\documentclass{WileyMSP-template}
\usepackage{amsmath}
\usepackage{booktabs}
\usepackage{listings}
\usepackage[utf8]{inputenc}
\usepackage{textcomp}
\usepackage{xcolor}
\usepackage[T1]{fontenc}
\usepackage{hyperref}
\usepackage[numbers, sort&compress]{natbib}
\usepackage{newunicodechar}
\newunicodechar{−}{\textminus}

\begin{document}

\pagestyle{fancy}
\rhead{\includegraphics[width=2.5cm]{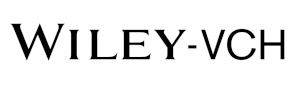}}

\title{Communicative Efficiency of Single vs. Multi-Axis Robot Neck Motion}

\maketitle


\author{Chapa Sirithunge$^1$*}
\author{Haewon Jeong$^2$}
\author{Qinghua Guan$^2$}
\author{Fumiya Iida$^{3}$}
\author{Josie Hughes$^2$}



\begin{affiliations}
$^1$University of Bristol\\
School of Engineering Mathematics and Technology, Tankard's Close, Bristol BS8 1TW, UK\\
Email Address: chapa.sirithunge@bristol.ac.uk*

$^2$EPFL\\
Institute of Mechanical Engineering, CH-1015 Lausanne, Switzerland

$^3$University of Tokyo\\
Department of Precision Engineering, Tokyo 113-8656, Japan

\end{affiliations}



\begin{abstract}
Nonverbal communication through head and neck movement is fundamental to human social signalling, yet how robotic neck morphology translates motion into communicative information remains poorly understood. We present an information-theoretic framework characterising robot neck movement as a communication channel, quantifying information transmitted and energy expended across varied configurations. Using a robotic neck platform, we recorded 84 video stimuli spanning three rotational degrees of freedom (DoF), varying amplitude, acceleration, and frequency, measuring Shannon entropy of pixel-change signals alongside energy consumption. A perceptual study validated communicative interpretations of each motion. While humans typically engage one axis per gesture, robots are unconstrained by biological architecture, motivating tests up to 3 DoF. Yet communicative information peaks at two DoF and decreases at three despite rising energy cost, a phenomenon we term the morphological information bottleneck. Motion parameter effects were parameter-dependent, some additive, others non-linear. We introduce the Motor Information Space, a framework mapping entropy against energy to expose communicative efficiency across morphologies, in which the optimal configuration achieves 5.26 bits at competitive energy cost. Perception data further confirm multi-axis movements reduce clarity. These findings challenge the assumption that anatomical completeness improves robotic expressiveness, establishing a quantitative basis for morphological design in robots, especially humanoids.

\end{abstract}

\section{Introduction}

The robot market is growing rapidly, and is expected to reach \$160 billion by 2030. Within this, the humanoid robot market predicted to have the highest growth rate of over 50\% by 2035 \cite{li2026comprehensive}. 
This has intensified the demand for design guidelines for such human-form factor robots, to consider not only the physical capabilities but also to address the capacity of social communication ~\cite{lassig2021robotics}. As humanoid platforms including Boston Dynamics Atlas, Agility Robotics Digit, and Figure 01 move from controlled laboratory settings into human-populated environments, the question of how their actions and morphology shapes social perception becomes increasingly consequential. However existing design frameworks largely address locomotion, manipulation, structural efficiency, and learning, leaving the communicative properties of individual body segments and in particular the neck underspecified.

Nonverbal communication through head movement and gaze constitutes a fundamental channel of social signalling in human interaction. Gestures and body movements function as structured communicative acts analogous to language, encoding intent, attention, and affective state through spatial and temporal patterning~\cite{kendon2004gesture}. This principle extends naturally to human-robot interaction, where Mutlu et al.~\cite{mutlu2009nonverbal} demonstrated that robot body movements unintentionally communicate information to human observers even when not explicitly designed to do so, and that small motion patterns reliably influence perceived attention and intention. Robot body orientation in particular has been shown to modulate human interpretation of robot behavior, with people attributing social meaning to motion cues in the absence of explicit communicative intent \cite{sirithunge2019proactive}.

Eyebrow movements \cite{homke2025eyebrow} are found to be functional communicative signals and mouth synchrony can significantly improve human-robot interaction \cite{burgos2012engaging}. Gaze direction and head orientation represent the most studied modalities within this broader space of robot nonverbal behavior. He et al.~\cite{he2024robot} demonstrated that robot gaze behaviors during autonomous navigation significantly affect perceived social presence, confirming that even task-directed head motion carries social meaning for human observers. At the control level, Haefflinger et al.~\cite{haefflinger2025data} proposed a data-driven controller for coordinated eye and head movements in triadic human-robot interactions, highlighting the complexity of generating contextually appropriate gaze behavior in multi-person settings. The technical landscape for enabling such behaviors has been systematically reviewed by Neuber and Culemann~\cite{neuber2025eyes}, who catalogued approaches for humanoid gaze-following, and by Fischer-Janzen et al.~\cite{fischer2024scoping}, who surveyed eye tracking-based control methods for assistive robotic systems. Complementary active perception strategies have been developed by Luzio et al.~\cite{luzio2024semantic} for foveal visual search, and improved eye-gaze tracking hardware has been proposed by Bandi and Thomas~\cite{bandi2023new} for real-time robotic applications. Field evidence for the significance of gaze in naturalistic encounters is provided by Zeng et al.~\cite{zeng2026encountering}, whose real-world eye-tracking study demonstrated structured attention patterns during human-robot encounters.

Despite this body of work, the communicative properties of neck movement as an independent expressive modality and its relationship to energy expenditure remain underexplored. 
Studies on human preferences for robot design suggest that perceptual cues including eyes and head orientation are prioritised by users even when other aspects of appearance are considered less important~\cite{sirithunge2023we, sirithunge2019situation}, underscoring the communicative significance of head and neck motion in shaping social perception during HRI. This gap is particularly consequential for humanoid design, where the neck can be readily actuated and the decisions affect mechanical properties and social legibility of head-based nonverbal communication. 
The present work addresses this gap by providing the first information-theoretic characterisation of robot neck movements as a communication channel, offering quantitative design guidelines directly applicable to next-generation humanoid platforms. Here we consider single axis motions (humanlike) such as roll, pitch yaw and multi-axis motions (beyond human) by combining these three basic movements. The proposed approach can be naturally extended to the kinematic design of robotic necks, building on prior work in this area \cite{ouerfelli2002kinematic, al2025humanoid, rajruangrabin2011robot}. Furthermore, recent studies on human–robot synchrony \cite{hu2026learning} and medical facial-expression simulation \cite{lalitharatne2021morphface} underscore the increasing interest in the physicality of robotic faces, particularly their embodied expressive and interactive capabilities. Shannon entropy has been used to quantify the complexity of facial expression signals generated by human avatars at different stages of transmission \cite{jack2014dynamic}, and this concept can be extended to analyse other signals transmitted by robotic systems such as necks.

\begin{figure}[!t]
    \centering
    \includegraphics[width=0.65\linewidth]{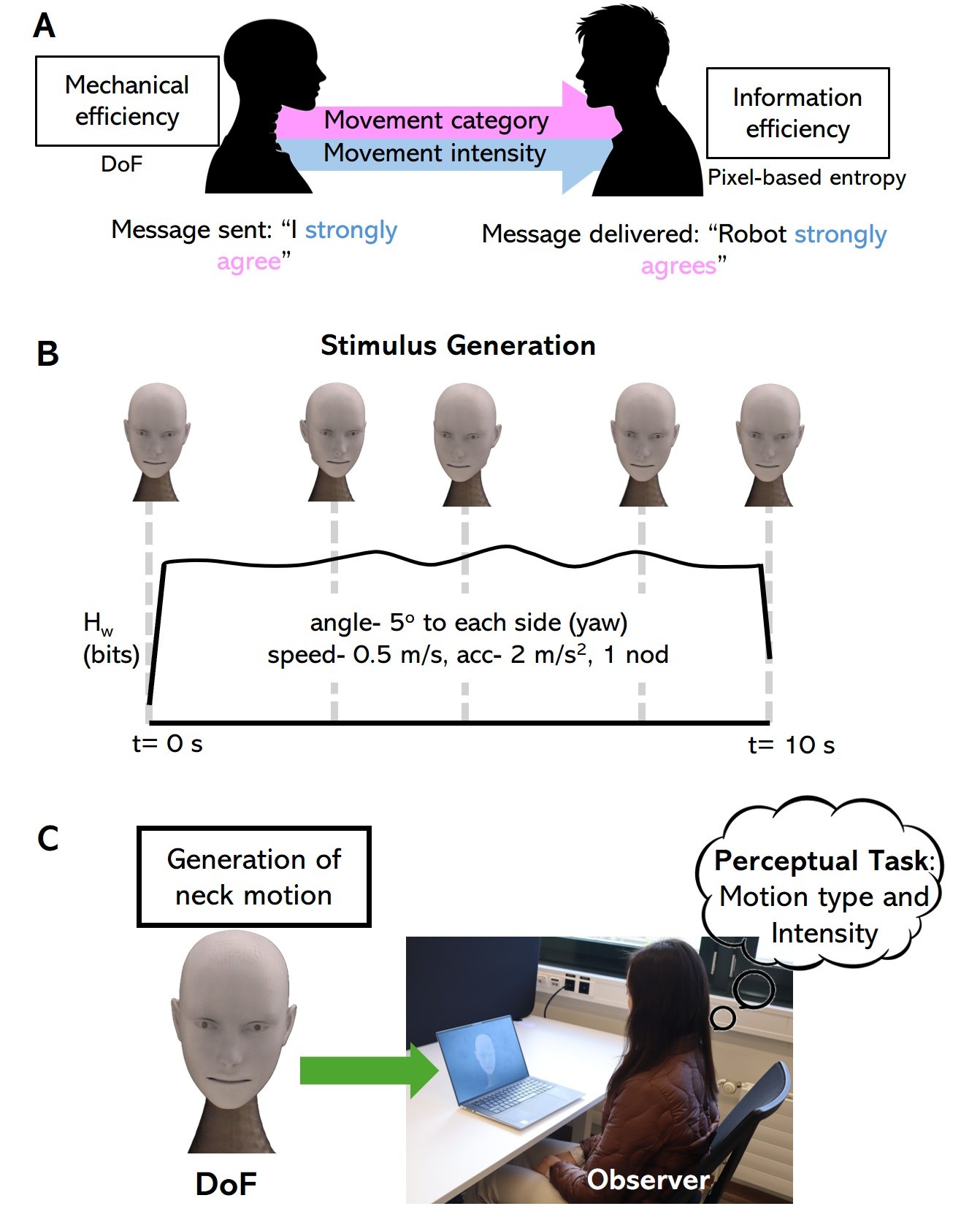}
    \caption{Neck communication framework and information-theoretic quantification of robotic neck motion
A) To convey an emotional message such as “I strongly agree,” a person may use a rapid sequence of head nods, either independently or alongside speech. In this sense, neck motion functions as an additional communication channel complementing verbal output. The sender (the robot) encodes the intended message through a specific movement strategy. Effective communication depends on shared expectations between sender and receiver; therefore, the receiver must accurately decode both the type and intensity of motion to interpret the message correctly. The mechanical output is characterised by the energy expended across the active degrees of freedom of the robotic neck, while the information received at the human observer's end is quantified by applying Shannon entropy to the distribution of mean absolute pixel differences between consecutive video frames \cite{mentzelopoulos2004key}. B) The mechanical characteristics of neck motion include the type of movement (roll, pitch, yaw), angular displacement, velocity, acceleration, and temporal features such as frequency or number of repetitions (nods). These parameters are mapped to a quantitative representation using principles from Shannon's information theory, enabling an information-theoretic description of motion. 
C) Robot-generated neck motions were recorded and presented to human observers in a perceptual study. This allowed us to establish a mapping between the mechanical parameters of motion generation and their perceived meaning, linking physical movement features to human interpretation.}
    \label{concept}
\end{figure}

The concept presented in this work is illustrated in Figure \ref{concept}. Robot neck movement is framed as a communication channel, characterised along two complementary axes. The first is a motor analysis examining the mechanical properties of communicative neck movements; roll, pitch, and yaw (motion categories) across varying motion parameters including speed, acceleration, and frequency (Figure \ref{concept}A). The second is an information-theoretic analysis quantifying the communicative content of each movement using Shannon entropy computed from the resulting motion stimuli (Figure \ref{concept}B). Together these two analyses form a design framework for robot necks that balances mechanical efficiency with maximum information transmission. To validate the perceptual validity of the generated motion features, the movements were presented to human participants in a structured survey, where participants rated their ability to perceive the intended communicative meaning and the intensity of each neck movement (Figure \ref{concept}C). This work seeks to advance understanding of two fundamental challenges: bridging the gap between physical efficiency and socially expressive behavior, and identifying how robot embodiments that depart from human features can nevertheless support intuitive and socially meaningful human–robot interaction.

\section{Methods}

\begin{figure}[!t]
    \centering
    \includegraphics[width=0.99\linewidth]{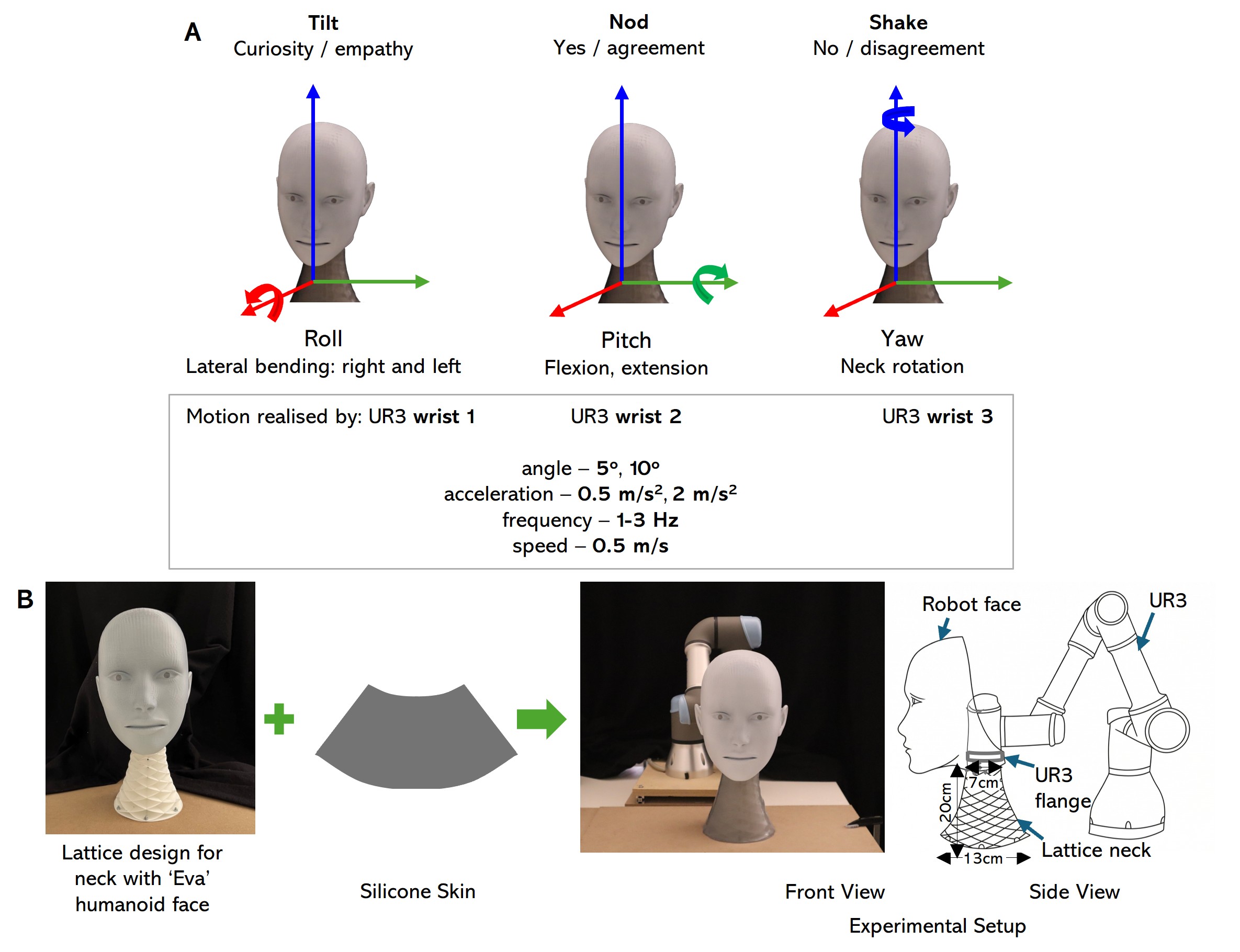}
    \caption{Experimental Protocol A) Neck motion design: The study considers three primary categories of neck movement: roll, pitch, and yaw, along with their commonly perceived meanings. The motion parameters include angular displacement, speed, acceleration, and temporal characteristics such as frequency or number of nods. These movements are physically realized using the joints of the UR3 robotic arm, with each joint corresponding to one of the rotational degrees of freedom. Each neck motion had two magnitudes: 5$^\circ$ and 10$^\circ$, two accelerations: 0.5 and 2 m/s$^2$ and three frequencies: once, twice and thrice. The speed was held constant at 0.5 m/s. B) Experimental setup: The setup consists of a robotic head and neck mounted on a UR3 manipulator. The upper end of the neck is attached to the UR3 robot as an end effector, while the lower end is fixed to the table. The neck is constructed using a lattice structure to enable a wide range of motion patterns and is covered with colored silicone skin to achieve a more natural appearance. The UR3 is configured such that wrist joints 1, 2, and 3 correspond to roll, pitch, and yaw motions respectively.}
    \label{exp}
\end{figure}
\subsection{Robot Neck Mechanism}
Rather than designing and fabricating a dedicated multi-motor neck mechanism, this study exploits the existing six-DoF architecture of the UR3  robotic arm to directly realize neck motion (Figure \ref{exp}). The compliant lattice neck structure, covered with coloured silicone skin to achieve a more natural appearance, is mounted such that its upper end is attached to the UR3 as an end effector, while its lower end is fixed to the table, with the robot's wrist joints driving the neck rather than the neck containing any actuation of its own.

This design choice offers several advantages over a purpose-built neck mechanism. First, it eliminates the need to engineer independent torque, velocity, and current sensing for each rotational axis. Second, it allows precise, repeatable, and independently controllable motion parameters: each of the three primary neck motions, roll, pitch, and yaw, is mapped directly onto wrist joints 1, 2, and 3 of the UR3 respectively, allowing any combination of axes to be activated individually or simultaneously through standard robot motion commands. Third, it enables systematic parametric variation of motion characteristics that would be difficult to achieve with a custom mechanism: each neck motion was specified with two angular magnitudes (5$^\circ$ and 10$^\circ$), two accelerations (0.5 and 2 m/s$^2$), and three repetition frequencies (one, two, or three nods), while speed was held constant at 0.5 m/s, all governed directly through the UR3's native motion planner without requiring custom firmware. This  substantially reduces development time and mechanical complexity while providing sensing accuracy. More details about this platform is given in Section \ref{exp_setting}.

\subsection{Motion Analyses}

To analyse the efficiency of neck-based motion, we adopted two complementary approaches. The first focused on the mechanical efficiency of different types of neck movements, evaluating them in terms of degrees of freedom (DoFs) and energy consumption. This is discussed below under ``motor-based analysis''. The second approach used a ``pixel-based analysis'' to estimate the amount of information transferred across different video frames of neck movements during a single gesture. This method calculates information complexity by measuring pixel intensity changes between consecutive video frames of the neck while keeping the camera angle and position fixed. Shannon entropy is then applied to the distribution of these pixel changes to quantify how much motion variability and visual complexity are present in the neck region. Changes in the skin around the neck resulting from these movements were also considered, as they contributed to the overall amount of information conveyed through neck motion. Similarly, shadows created during the movements were included, as they also affected the visual transfer of information.

\subsection{Motor-based Analysis}
\label{motor}

To quantify neck movements, we measure the angular displacement associated with three fundamental motion components: flexion– extension (nodding the head forward and backward/pitch), lateral bending (tilting the head toward the left or right shoulder/roll), and axial rotation \cite{doss2023comprehensive, bland1990anatomy}. These movements correspond to the primary degrees of freedom of the human cervical spine and form the basis of head orientation during communication and perception. In many human communicative behaviors such as indicating attention, acknowledging information, or directing gaze, only a subset of these motions is typically employed at a given moment to convey directional or attentional cues. By quantifying these angular motions, the proposed framework provides a principled way to estimate how much directional information a robotic neck can generate during an interaction.

The three robot neck movements used in this study and their communicative interpretations: tilt (roll), nod (pitch), and shake (yaw), correspond to curiosity, agreement, and disagreement respectively \cite{mower2010framework, ishi2010head, poggi2007mind}. Parametric values used to realise the movements and their corresponding interpretations used in this study are shown in Figure \ref{exp}A.

Key parameters used here are as follows.
\begin{itemize}
    \item Motion types: roll, pitch, yaw, roll+pitch, roll+yaw, 
    pitch+yaw, roll+pitch+yaw
    \item Angles of each motion: $\theta \in \{5^{\circ}, 10^{\circ}\}$
    \item DoF levels: $n_{\text{DoF}} \in \{1, 2, 3\}$
\end{itemize}

Single-axis motions (roll, pitch, yaw) can be combined to produce multi-axis motion; for example, simultaneous roll and pitch yields a 2-DoF movement beyond typical human capability. Each axis may be assigned different amplitudes, such as $5^{\circ}$ and $10^{\circ}$.

Key assumptions used in this context are below. 

Electrical energy reported 
by the UR3 controller is treated as a valid proxy for mechanical 
energy expenditure. Joint-level contributions are assumed 
independent and additive. A polynomial trendline of degree 
$d = 2$ is fitted per angle condition across motion types to 
visualise trends.

 To estimate the total energy consumed by each neck movement, we first calculate the torque produced at each joint, from which instantaneous mechanical power and cumulative energy can subsequently be derived. \textbf{Motor output-shaft torque} for joint $i$ was estimated from the measured
current using the linear motor model
 
\begin{equation}
  \tau_i(k) = K_{t,i} \cdot I_i(k),
  \label{eq:torque}
\end{equation}
 
\noindent
where $K_{t,i}$ is the motor torque constant for joint $i$ [Nm\,A$^{-1}$] and k is the discrete time step index.
The numerical values used in the experiment are listed in Table~\ref{tab:kt} under Section \ref{sup_energy}.  These constants represent
the combined effect of the motor's back-EMF constant and gearbox efficiency,
mapping measured current directly to output-shaft torque; no separate gear
ratio term is required.

 
The \textbf{instantaneous mechanical power} drawn by joint $i$ at time step $k$ is
 
\begin{equation}
  P_i(k) = \bigl|\tau_i(k) \cdot \omega_i(k)\bigr|,
  \label{eq:power}
\end{equation}
 
where $\omega_i(k)$ is the instantaneous angular velocity of joint $i$
at time step $k$ measured in radians per second (rad/s).  The absolute value ensures that regenerative phases (negative
$\tau_i \omega_i$) contribute positively to the energy budget rather than
cancelling the energy consumed during the acceleration phase.The total
instantaneous power across all joints is
 
\begin{equation}
  P_{\text{total}}(k) = \sum_{i=0}^{5} P_i(k).
  \label{eq:ptotal}
\end{equation}
 
 
With instantaneous power available at each time step, the total energy consumed by the movement can be obtained by integrating $P_{\text{total}}(k)$
over time. \textbf{Cumulative energy}, $E(k)$ throughout the movement was computed by numerical integration of
$P_{\text{total}}(k)$ using the trapezoidal rule:
 
\begin{equation}
  E(k) = E(k-1) + \frac{1}{2}
         \Bigl[P_{\text{total}}(k-1) + P_{\text{total}}(k)\Bigr] \cdot \Delta t_k,
  \label{eq:energy}
\end{equation}
 
where $\Delta t_k = t_k - t_{k-1}$ is the actual elapsed time between
consecutive samples (used instead of the nominal $\Delta t$ to absorb
scheduling jitter). The initial condition is $E(0) = 0$.


All \textbf{energy} and \textbf{power} metrics were obtained from the UR3 robot 
controller, which reports instantaneous joint torques 
$\tau_i(t)$ and joint velocities $\dot{q}_i(t)$ for each 
active joint $i$. The \textbf{instantaneous mechanical power} is 
computed as:

\begin{equation}
    P(t) = \sum_{i=1}^{n} \tau_i(t) \cdot \dot{q}_i(t)
    \label{eq:power}
\end{equation}

\noindent where $n$ is the number of active joints corresponding 
to the DoF configuration. Total energy consumption over a 
movement of duration $T$ is computed as:

\begin{equation}
    E = \int_0^T P(t)\, dt \approx \sum_{t=1}^{N} P_t \cdot \Delta t
    \label{eq:energy}
\end{equation}

\noindent where $\Delta t$ is the controller sampling interval 
and $N$ is the total number of samples. Mean and peak power 
are defined as:

\begin{equation}
    \bar{P} = \frac{1}{T} \int_0^T P(t)\, dt 
    \approx \frac{1}{N} \sum_{t=1}^{N} P_t
    \label{eq:mean_power}
\end{equation}

\begin{equation}
    P_{\text{peak}} = \max_{t \in [0,\,T]} P(t)
    \label{eq:peak_power}
\end{equation}

More details of the calculations and derivations can be found under supplementary information in section \ref{sup_energy}.

\subsection{Pixel-based Shannon Entropy as a Measure of Communicative Information}
\label{sec_pixel}

In addition to motor signal analysis, an image-based method is used to estimate the amount of visual information generated by neck movements. A video sequence of the robotic head performing controlled neck motions is recorded using a fixed camera. The face and neck region is extracted from each frame to isolate the motion of interest.

The information content of each neck motion was quantified using Shannon entropy computed from a pixel-level motion signal extracted from video recordings. For each video, consecutive grayscale frames were compared by computing the mean absolute pixel difference between frame t and frame t-1, producing a one-dimensional time series that captures the magnitude of visual change at each moment in time. This signal serves as a proxy for the perceptual salience of the movement; a larger pixel difference indicates a more visually distinct motion, while a smaller difference indicates a more uniform or predictable one. Shannon entropy $H$ as the dependent variable, computed from the pixel-change signal $\delta_t$ extracted from 84 video 
recordings of the robot neck across combinations of roll, pitch, 
and yaw axes. The pixel-change signal is defined as:

\begin{equation}
    \delta_t = \frac{1}{WH} \sum_{x,y} \left| I_t(x,y) - I_{t-1}(x,y) \right|
    \label{pixel}
\end{equation}

where $ \delta_t$ is the mean absolute pixel difference at time step $t$, representing the average magnitude of visual change between consecutive frames. This is the pixel-change signal used as the information-bearing stimulus. $t$ is the discrete frame index, running from $t = 2$ to $t = N$ where $N$ is the total number of frames in the video. $W$ is the width of the video frame in pixels and $H$ the height of the video frame in pixels. The product $WH$ is therefore the total number of pixels per frame, used to normalise the sum into a mean value.
$I_t(x,y)$ is the grayscale intensity of the pixel at spatial coordinates $(x,y)$ in frame $t$, taking values in the range [0, 255]. 
$I_{t-1}(x,y)$ is the grayscale intensity of the pixel at the same spatial coordinates $(x,y)$ in the preceding frame $t-1$.

Shannon entropy was then computed over this signal by constructing a 64-bin histogram of the pixel difference values across the entire video, normalising the bin counts to obtain a probability distribution, and applying the standard entropy formula:

\begin{equation}
    H = -\sum_{i\,:\,p_i > 0} p_i \log_2 p_i, 
    \quad 
    p_i = \frac{c_i}{\sum_{j\,:\,c_j>0} c_j}
    \label{H}
\end{equation}

where $p_i$ is the probability mass in bin $i$. The resulting entropy value $H$, measured in bits, quantifies the unpredictability of the pixel change distribution,  higher entropy indicates a more varied and informationally rich movement, while a lower entropy indicates a more repetitive or predictable signal. $i$ is the bin index of the histogram, running over all non-empty bins only, as indicated by the condition $i: p_i > 0$. Empty bins are excluded to avoid undefined $\log_2(0)$ terms. $p_i$ is the normalised probability mass assigned to bin $i$, representing the proportion of pixel-change signal samples that fall within that bin's range. $c_i$ is the raw count of pixel-change signal samples $\delta_t$ falling within bin $i$ of the histogram. $j$ is a summation index running over all non-empty bins, used in the denominator of $p_i$ to normalise the raw counts into a valid probability distribution.
$\sum_{j\,:\,c_j>0} c_j$ is  the total number of samples across all non-empty bins, equivalent to the total number of frames minus one in the video, used as the normalisation factor.
$\log_2$ is logarithm base 2, which ensures entropy is expressed in bits. Results are constructed using B = 64 uniform bins spanning the observed range of $\delta_t$ for each individual video.

This approach was chosen because it requires no joint angle extraction, no marker-based tracking, and no prior knowledge of the robot's kinematics. The camera observes the robot from a fixed position against a static background, meaning all pixel change originates from neck movement alone while the background was fixed. The method therefore measures the information content of the movement as it would appear to an external observer, which is precisely the quantity of interest in a communication context.

Having obtained the entropy $H$ for each movement, we next combine it with the energy expended to produce that movement to evaluate communicative efficiency. Energy consumption was measured in parallel from the UR3 robot controller using the procedure described in Section~\ref{motor}, which reports total electrical energy draw $E$ in Joules according to Eq.~\ref{eq:energy}. The two measurements, entropy
$H$ (Eq.\ref{H}) and energy $E$, were then combined to compute the Entropic-to-Neuro-mechanical Information Efficiency (ENIE), defined as the ratio of communicative information to mechanical cost:

\begin{equation}
    \text{ENIE} = \frac{H}{E}
\end{equation}

Higher entropy values correspond to greater visual variability in the motion, indicating that the neck movement produces more distinguishable visual states. This method provides an estimate of the observable information generated by neck motion from a visual perception perspective. Further details on the implementation of these equations in the hardware setup are provided in Section \ref{Pixel_plots}.
 
\subsection{Assessment of Human Information Reception: A Survey}

 To evaluate how much information neck movements convey to human observers, a human perception study was conducted. Participants were shown short video clips of the robotic head performing neck movements while maintaining a neutral facial expression.  Video length ranged from 6-16 s based on the number of nods. This idea is demonstrated in Figure \ref{concept}C.

Participants were asked to evaluate how informative the movement appears in terms of conveying the direction or focus of the robot’s attention. Responses were recorded using a predefined rating scale that measures perceived informativeness or clarity of the motion. Participants were recruited to provide independent evaluations of the recorded movements. Once the video was played, two questions were presented to the participant:

\begin{itemize}
    \item Q1. What does this head motion of robot convey to you? 
\textit{    Options to select: Agreement/Yes, Disagreement/No, Curiosity, Nothing, }Something else
    \item Q2. From a scale of 0 to 5, what is the intensity of that nonverbal message?
\textit{(0 – Null, 3 – Appropriate, 5 – Fully informative)}
\end{itemize}

The responses are aggregated to obtain a distribution of perceived information values across participants. This distribution was then be analyzed statistically or converted into an entropy-based measure to estimate the human-perceived information content of the neck movement. This evaluation complements the motor-signal and image-based analyses by capturing the communicative effectiveness of neck motion during human–robot interaction.

\subsection*{Experimental Setting}
\label{exp_setting}

The hardware platform employed in this study is Eva, a humanoid robot face and an open-source system  to investigate human–robot communication \cite{faraj2021facially}. We utilized only the face cover design of Eva, without using the original hardware associated with it. 

To achieve human-like neck behaviors, we adopt a trimmed helicoid structure as the compliant core of the neck \cite{guan2023trimmed}. Going beyond simply adjusting the trimming area (i.e., retaining more material around the central axis), we further tailor the cross‑sectional geometry and the distribution of cell parameters (such as helical angle) along the axial direction. Inspired by the concept of programmable geometric topology \cite{guan2025lattice}, these axial gradients allow the helicoid to be locally tuned: the neck remains flexible in bending for natural lateral movements while maintaining sufficient axial stiffness to resist necessary compression, thereby ensuring structural stability. The resulting mechanical anisotropy closely mimics the biomechanical response of the human neck. 

A Universal Robots UR3  6-DoF  manipulator was used to create neck movements. Flange of UR3 was fixed on to one end of the neck. UR3 was connected to a host PC via a 100\,Mbit/s Ethernet link. All data acquisition and motion commands were issued through the Real-Time Data Exchange (RTDE) interface using the \texttt{ur\_rtde} Python library, which provides access to the robot controller's internal state at the native control loop rate of 125\,Hz. At each sample, joint currents and velocities are retrieved, from which per-joint torque, mechanical power, and cumulative energy consumption are computed and logged to CSV files for subsequent analysis.

Bottom end of the neck was fixed on to a plain surface. Face was attached to the flange to move with the tool (neck). A Silicone skin (Ecoflex 00-10) was attached around the neck to make it look more real. During the experiment, UR3 was covered with a black background. This setup is shown in Figure \ref{exp}B. 

\section{Results}

This section presents the results of both the motor-based and pixel-based analyses, along with the observations from the human assessment.

\subsection{Motor-based Analysis}

\begin{figure}[!b]
    \centering
    \includegraphics[width=0.99\linewidth]{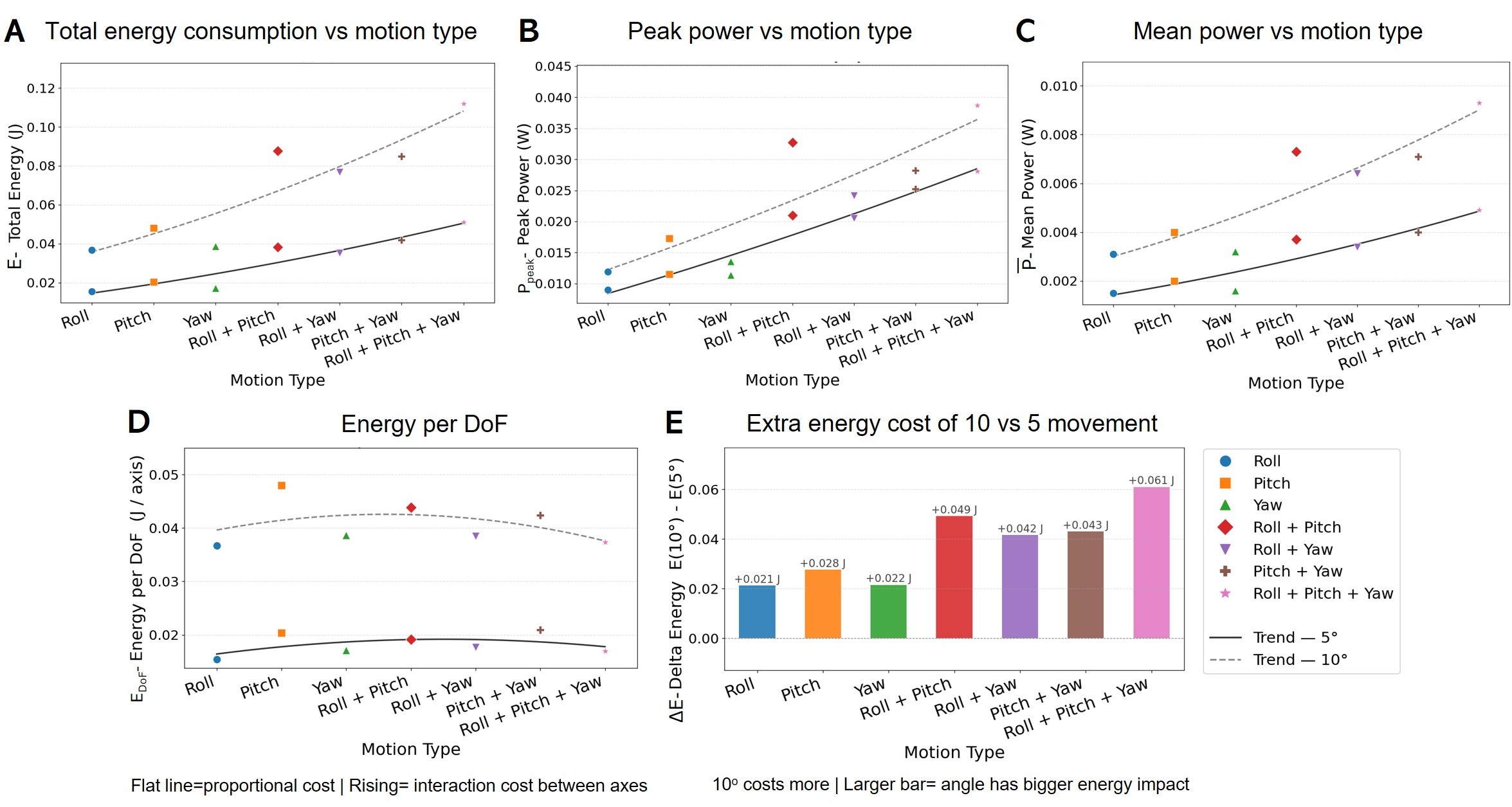}
    \caption{All results related to energy consumption during different motions A) Total energy consumption calculated according to eq.~\ref{eq:energy} for different motion strategies B-C) Peak and mean power consumption plotted against motion strategies D) Energy spent during the implementation of each motion  plotted against DoF. E) Energy cost between $10^\circ$ and $5^\circ$ degrees neck movements. Legend for all graphs shown on bottom right.   }
    \label{energy_plots_1}
\end{figure}

Results related to the motor-based analysis are shown in Figure \ref{energy_plots_1}. 





When the total energy consumption is compared against motion type (see Figure~\ref{energy_plots_1}A), energy increases monotonically from single-axis to three-axis combinations, which is expected. However, the increase is not linear where the $10^\circ$ trendline (dashed) rises much more steeply than the $5^\circ$ line (solid), meaning larger angles disproportionately amplify energy costs as more axes are added. Within single axes, Yaw is notably cheaper than Roll and Pitch at both angles, suggesting the yaw joint requires less torque for the same angular displacement. Roll + Pitch stands out as an outlier and its $10^\circ$ energy ($\sim 0.086$ J) is higher than Roll + Yaw and Pitch + Yaw, suggesting the roll--pitch combination has a specific mechanical interaction cost. The full 3-DoF combination at $10^\circ$ ($\sim 0.115$ J) is roughly 6$\times$ more expensive than single-axis Roll at $5^\circ$ ($\sim 0.018$ J).

The pattern between peak power and motion type mirrors total energy, but the spread is larger, meaning peak power grows faster than average power as DoF increases. This is shown in Figure~\ref{energy_plots_1}B. Roll + Pitch again stands out as the highest-cost 2-DoF combination at $10^\circ$ ($\sim 0.033$ W), exceeding Roll + Yaw and Pitch + Yaw noticeably. This suggests Roll + Pitch produces the most impulsive loading where the two axes likely create mechanical coupling that forces brief high-torque events. The $5^\circ$ trendline is much flatter than $10^\circ$, confirming that at small angles the peak power demand is relatively insensitive to axis combination. This has a practical implication: if your robot has a peak power budget, $10^\circ$ movements in combined axes may exceed it even if mean power stays within limits. These findings can be applied to robots of different scales and morphologies.

Mean power shown in Figure~\ref{energy_plots_1}C follows the same trend as total energy, but at a much smaller scale (milliwatts range). The gap between $5^\circ$ and $10^\circ$ is proportionally consistent across motion types, unlike peak power where the gap widens to a certain degree. This confirms that the extra cost of larger angles is primarily impulsive (captured in peak power) rather than sustained. The near-flat $5^\circ$ trendline is particularly notable. At $5^\circ$, mean power is almost independent of the number of axes, meaning the sustained electrical load of the robot barely changes whether it moves one axis or three at small angles.

Based on Figure~\ref{energy_plots_1}D, the $5^\circ$ trendline (solid) is nearly flat at $\sim 0.016$--$0.020$ J/axis across all combinations, meaning at small angles, each axis contributes roughly equal energy regardless of whether it is acting alone or combined. This is the ideal proportional scaling. The $10^\circ$ trendline (dashed) is curved and higher, peaking around Roll + Pitch and then declining slightly toward Roll + Pitch + Yaw. Pitch is a clear outlier at $10^\circ$ ($\sim 0.048$ J/axis), far above the trend, suggesting the pitch joint specifically incurs a disproportionate energy cost at larger angles, possibly due to its mechanical configuration fighting gravity. The fact that Roll + Pitch + Yaw brings the per-axis cost back down slightly suggests the third axis (yaw) is relatively cheap and effectively dilutes the combined cost. The practical conclusion is that the proportional energy model holds well at $5^\circ$ but breaks down at $10^\circ$, with Pitch being the primary driver of non-linearity.

A bar chart is used to visualise $\Delta E$ per motion type, with a zero reference line indicating conditions where amplitude has no net energy cost. Positive values indicate that the $10^{\circ}$ condition requires greater energy expenditure. This plot directly quantifies the sensitivity of each motion strategy to movement amplitude and reveals whether the angle effect is additive or exhibits 
interaction across axes. Figure~\ref{energy_plots_1}E plots the extra energy cost across different motion types. Defining delta energy as $\Delta E = E(10^\circ) - E(5^\circ)$, all bars are positive where $10^\circ$ always costs more than $5^\circ$, which is expected. However, the magnitude varies substantially and non-uniformly. Single axes cost an extra $\sim 0.021$--$0.028$ J for the angle increase, but combined axes cost significantly more: Roll + Pitch adds 0.049 J, Roll + Yaw 0.042 J, Pitch + Yaw 0.043 J, and Roll + Pitch + Yaw 0.061 J. The jump from single to combined axes is not purely additive.  If $\Delta E$ were additive, Roll + Pitch would be approximately $0.021 + 0.028 = 0.049$ J, which matches almost exactly. Roll + Yaw should be $0.021 + 0.022 = 0.043$ J but is 0.042 J, also near-additive. This is a strong result: the extra energy cost of larger angles is approximately additive across axes, meaning there is no significant interaction between axes in terms of how angle affects energy. The full 3-DoF combination (0.061 J) is slightly less than the sum of its parts ($0.021 + 0.028 + 0.022 = 0.071$ J), suggesting minor sub-additive interaction at 3 DoF, the robot may be sharing some mechanical work across joints when all three move together.

More details of the calculations related to the above plots can be found under supplementary information in section \ref{sup_energy}.

\subsection{Pixel-based Analysis}

\begin{figure}
    \centering
    \includegraphics[width=0.99\linewidth]{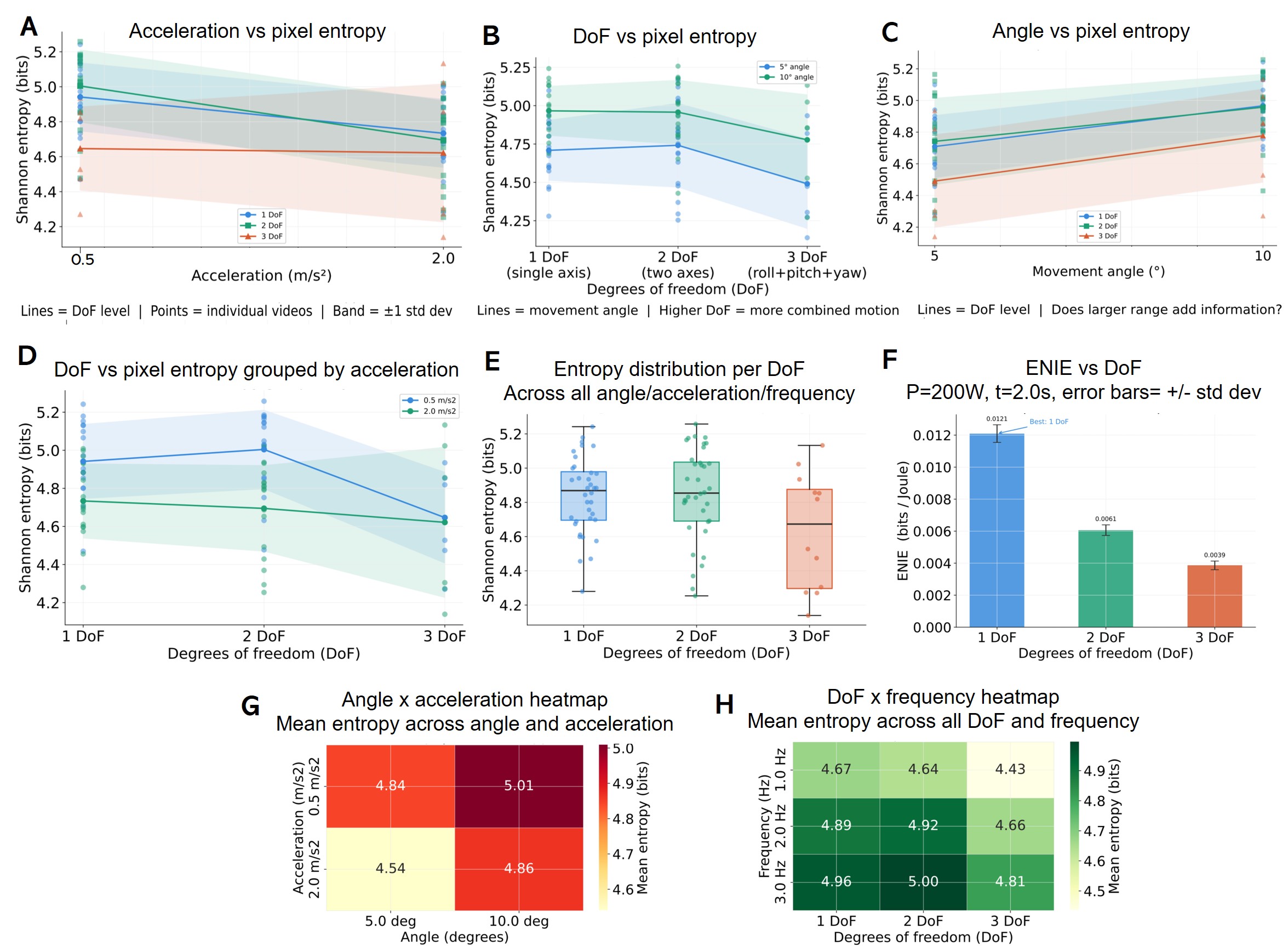}
    \caption{Pixel entropy analysis of robot neck movements across motion parameters. All entropy values were computed using Shannon entropy ($H$, bits) of the mean absolute pixel-change signal extracted from 84 video recordings of a UR3 robot neck across roll, pitch, and yaw combinations. (A) Acceleration vs pixel entropy: entropy decreases with increasing acceleration across all DoF levels, with 2 DoF consistently producing the highest entropy. Lines represent DoF level, points are individual videos, and bands represent $\pm 1$ standard deviation. (B) DoF vs pixel entropy: entropy peaks at 2 DoF and falls at 3 DoF for both $5^\circ$ and $10^\circ$ angle conditions, revealing the morphological information bottleneck. (C) Angle vs pixel entropy: larger movement amplitude produces consistently higher entropy across all DoF levels, with parallel lines confirming an additive, non-interacting effect. (D) DoF vs pixel entropy grouped by acceleration: slow acceleration (0.5 m/s$^2$) produces higher entropy than fast (2.0 m/s$^2$) across all DoF levels, with both conditions exhibiting the same 2 DoF peak. (E) Entropy distribution per DoF across all angle, acceleration, and frequency conditions: 1 DoF and 2 DoF share near-identical medians ($\sim$4.87 bits) while 3 DoF shows a lower median ($\sim$4.67 bits) and wider spread. (F) ENIE vs DoF at $P = 200$ W, $t = 2.0$ s: communicative efficiency drops monotonically from 0.0121 bits/J at 1 DoF to 0.0039 bits/J at 3 DoF, with 1 DoF identified as the most energy-efficient configuration. Error bars represent $\pm 1$ standard deviation. (G) Angle $\times$ acceleration heatmap: mean entropy peaks at $10^\circ$, 0.5 m/s$^2$ (5.01 bits) and is lowest at $5^\circ$, 2.0 m/s$^2$ (4.54 bits), with angle having a stronger effect than acceleration. (H) DoF $\times$ frequency heatmap: mean entropy is maximised at 2 DoF, 3 Hz (5.00 bits) and minimised at 3 DoF, 1 Hz (4.43 bits), confirming 2 DoF as the empirically optimal morphological configuration.}
    \label{Pixel_plot_2}
\end{figure}

Results related to the pixel-based analysis are shown in Figure \ref{Pixel_plot_2}.  All the equations used to calculated entropy are based on Section \ref{sec_pixel} and its derivations to create each plot are explained in Section \ref{Pixel_plots}. 

The most striking observation when comparing acceleration vs entropy,  is that entropy decreases as acceleration increases, which is counterintuitive  (see Figure \ref{Pixel_plot_2}A). The likely explanation is that high acceleration causes rapid, abrupt pixel changes that are actually more predictable in their distribution. A sharp spike followed by stillness is a peaky histogram, which is low entropy. Slow acceleration (0.5 m/s$^2$) produces smoother, more sustained pixel change that fills the distribution more evenly, giving higher entropy. This is an important result for the ENIE framework: \textbf{gentle movements are informationally richer than aggressive ones for a fixed angle}. It can be seen that 2 DoF consistently sits above 1 DoF, and 3 DoF sits below both across all acceleration levels; a pattern that recurs across all plots. 

In Figure \ref{Pixel_plot_2}B, entropy peaks at 2 DoF and drops at 3 DoF for both angle conditions. This holds for both 5$^\circ$ and 10$^\circ$ movements. The 3 DoF drop is substantial and this is roughly 0.2–0.25 bits below 2 DoF. This suggests that when all three axes move simultaneously, the pixel signal becomes more predictable rather than less, likely because the combined motion produces a more uniform blur or sweep across frames that reduces distributional variety. This directly supports the ENIE prediction that 2 DoF is the optimal morphological configuration, and gives  an empirical justification rather than just a theoretical one. In contrast, angle has a positive but modest effect (see Figure \ref{Pixel_plot_2}C). As all three DoF lines are parallel, confirming the angle effect is additive and does not interact with DoF. The 3 DoF line again sits lowest throughout.

According to Figure \ref{Pixel_plot_2}D, slow acceleration (0.5 m/s$^2$) produces consistently higher entropy than fast (2 m/s$^2$) across all DoF levels, reinforcing previous observations. The gap between the two lines is roughly 0.2 bits and remains stable across DoF, confirming acceleration and DoF have independent, additive effects rather than interacting. Figure \ref{Pixel_plot_2}E confirms the pattern: 1 DoF and 2 DoF have nearly identical median entropy (~4.87 bits), while 3 DoF has a notably lower median (~4.67 bits) and a wider spread (larger IQR). The wider spread at 3 DoF is also meaningful and the combined roll+pitch+yaw motion is more sensitive to the other parameters (angle, frequency, acceleration), producing more variable results. The 1 DoF box is also wide, suggesting individual axes vary considerably. The 2 DoF box is relatively compact and high, making it the most consistent and information-rich configuration.

ENIE drops monotonically (Figure \ref{Pixel_plot_2}F): 1 DoF = 0.0121, 2 DoF = 0.0061, 3 DoF = 0.0039 bits/Joule. Because energy scales with DoF (more joints active = more power drawn), and entropy does not increase proportionally, efficiency halves with each added DoF. This makes 1 DoF the most energy-efficient in absolute terms. However this needs careful interpretation where 1 DoF produces less total information even if it does so more efficiently per joule. The research implication is that the optimal design depends on the application: if total information capacity matters, 2 DoF wins; if energy budget is the primary constraint, 1 DoF wins.

According to angle$\times$acceleration heatmap (Figure \ref{Pixel_plot_2}G), the highest entropy condition is 10$^\circ$ + 0.5 m/s$^2$ = 5.01 bits and the lowest is 5$^\circ$ + 2.0 m/s$^2$ = 4.54 bits. The angle effect is larger than the acceleration effect in absolute terms. Going from 5$^\circ$ to 10$^\circ$ gains ~0.3 bits regardless of acceleration, while going from 2.0 to 0.5 m/s$^2$ gains ~0.15–0.17 bits. Angle is the more powerful driver of entropy than acceleration. There is no meaningful interaction visible and the two effects simply add. DoF × frequency heatmap (Figure \ref{Pixel_plot_2}H) shows that the global maximum is 2 DoF + 3 Hz = 5.00 bits, confirming this as the empirically optimal operating point. The pattern shows frequency is consistently beneficial while 3 DoF consistently underperforms. The worst condition is 3 DoF + 1 Hz = 4.43 bits. The 1 DoF and 2 DoF columns are nearly identical at every frequency (never more than 0.05 bits apart), meaning adding a second axis costs almost nothing informationally while halving ENIE, reinforcing the 1 DoF efficiency argument.

When all videos ranked by pixel entropy, the top-ranked video is roll+yaw | 10$^\circ$ | 0.5 m/s$^2$ | 3 Hz = 5.258 bits. Interestingly this is a 2 DoF combination, not 3 DoF. The top 10 entries are dominated by 2 DoF combinations (roll+yaw, pitch+yaw, roll+pitch) at $10^\circ$ and 0.5 m/s$^2$, confirming that the axis combination matters. Roll+yaw appears to produce the most visually distinct pixel signal, possibly because these two axes move in geometrically orthogonal visual planes as seen by the front-facing camera. The bottom of the ranking is dominated by 3 DoF at small angles, high acceleration, and low frequency and this is exactly the worst combination predicted by all previous plots.

It is worth noticing from these results that acceleration hurts entropy and that 3 DoF underperforms 2 DoF. Both challenge naive assumptions and both are grounded in a clear physical mechanism  can be articulate for future robotic designs concerning gestural complexity. 

\begin{figure}
    \centering
    \includegraphics[width=0.99\linewidth]{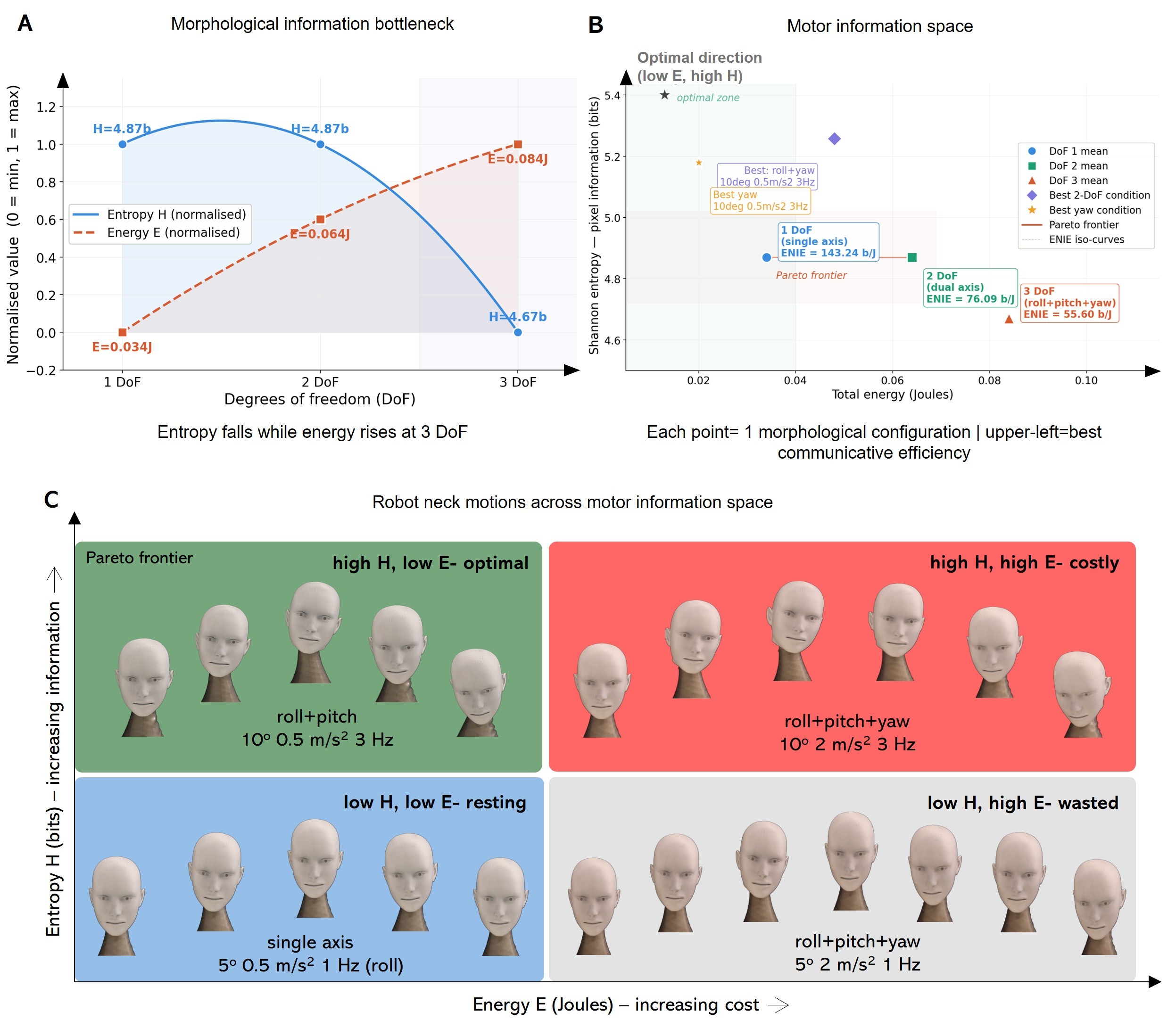}
    \caption{A) Morphological information bottleneck: Normalised Shannon entropy H and total energy E plotted against degrees of freedom (DoF). Entropy peaks at 2 DoF then falls at 3 DoF while energy rises monotonically, producing a divergence at 3 DoF where additional mechanical complexity yields diminishing informational return — the morphological information bottleneck. B) Motor Information Space (MIS): Each point represents a neck morphology configuration plotted by energy cost (x-axis) and pixel information entropy (y-axis). ENIE iso-curves (dashed) connect configurations of equal communicative efficiency. The Pareto frontier marks configurations non-dominated on both axes. Individual best conditions (diamond, star) demonstrate that parameter choice within a DoF level rivals morphology in determining communicative output. No 3 DoF configuration lies on or near the frontier. C) Robot neck postures mapped onto the Motor Information Space (MIS), defined by Shannon entropy H (pixel information, vertical axis) and total electrical energy E (horizontal axis). Dashed arcs indicate gaze direction distributions; arc spread reflects informational richness and base bars reflect energy expenditure. Top-left: optimal zone — varied gaze coverage at low energy cost, on the Pareto frontier. Top-right: high-cost expressive zone — wide arc coverage at disproportionate energy expense. Bottom-left: resting zone — minimal pixel variation, near-zero energy draw. Bottom-right: bottleneck zone — all three axes active yet gaze directions cluster redundantly, yielding low informational complexity despite high energy cost. No tested 3 DoF configuration lies on or near the Pareto frontier. The supplementary video presents the robot neck motion clips corresponding to the configurations depicted in this figure, illustrating their distribution across the motor information space. }
    \label{conclusion}
\end{figure}

Figure \ref{conclusion}A makes the decoupling between entropy and energy visually explicit. Both metrics are normalised to a 0-to-1 scale so they can be overlaid on the same axis despite having completely different units. A value of 1 means that configuration has the highest value of that metric, 0 means the lowest. The real values annotated on each data point (H=4.87b, H=4.87b, H=4.67b for entropy and 0.034J, 0.064J, 0.084J for energy) let  the absolute numbers be read even though the curves are normalised, so the plot communicates both the shape of the relationship and the actual magnitudes at once. The normalised entropy curve stays flat from 1 to 2 DoF (both configurations have the same entropy, so they overlap at the same normalised position), then drops at 3 DoF. The polynomial trendline shows this as a gentle inverted curve that peaks around 2 DoF and declines toward 3. The normalised energy curve rises steadily from left to right where more DoF always means more energy, with no plateau. The critical moment is at 3 DoF, marked by the faint purple shading. This is where the two curves visibly diverge and here the entropy goes down while energy goes up. Before 3 DoF the curves are moving in roughly compatible directions, but at 3 DoF they pull apart. This divergence is the \textbf{morphological information bottleneck}: the physical system has hit a point where adding more mechanical complexity no longer produces more communicative information, but still consumes more energy.

Figure \ref{conclusion}B places each DoF configuration as a single point in a 2D space where the Shannon entropy (bits of pixel information) is compared against energy consumed (Joules). The idea is borrowed from information theory's channel capacity framework and we are mapping out the entire communicative landscape of the robot neck in one image. The faint dashed lines running diagonally are ENIE iso-curves. Every point on a given iso-curve has the same bits-per-joule efficiency, similar to contour lines on a map represent the same altitude. Points in the upper-left corner are better on both dimensions simultaneously; more information for less energy. Points in the lower-right are worse on both. The Pareto frontier is the red line connecting the configurations that cannot be strictly beaten. A configuration is on the Pareto front if there is no other configuration that is both cheaper and more informative at the same time. In your data, only 1 DoF sits on the frontier because it delivers the same entropy as 2 DoF at lower energy. 3 DoF falls off the frontier entirely and it costs more energy and produces less information than both alternatives, making it dominated. The diamond marker shows the best individual video condition across your entire dataset (roll+yaw, 10$^\circ$, 0.5 m/s$^\circ$, 3 Hz at 5.26 bits), and the star shows the best single-axis condition (yaw at similar parameters). These individual best points sit well above the DoF mean points, showing that the choice of parameters within a configuration matters almost as much as the choice of DoF itself. The star marker in the top-left corner marks the optimal direction, to move towards lower energy and higher entropy simultaneously, which is the upper-left corner of this space.

\subsection{Survey}
\label{survey}


Seventeen participants took part in the study, comprising 8 females and 9 males, with a mean age of 30.8 years (SD = 3.8, range 25–39). A total of 714 valid response pairs were collected across 84 robotic neck-movement stimuli.  Participants were drawn from diverse backgrounds, originating from regions across Europe, Asia and the Americas.

\begin{figure}[!t]
    \centering
    \includegraphics[width=0.97\linewidth]{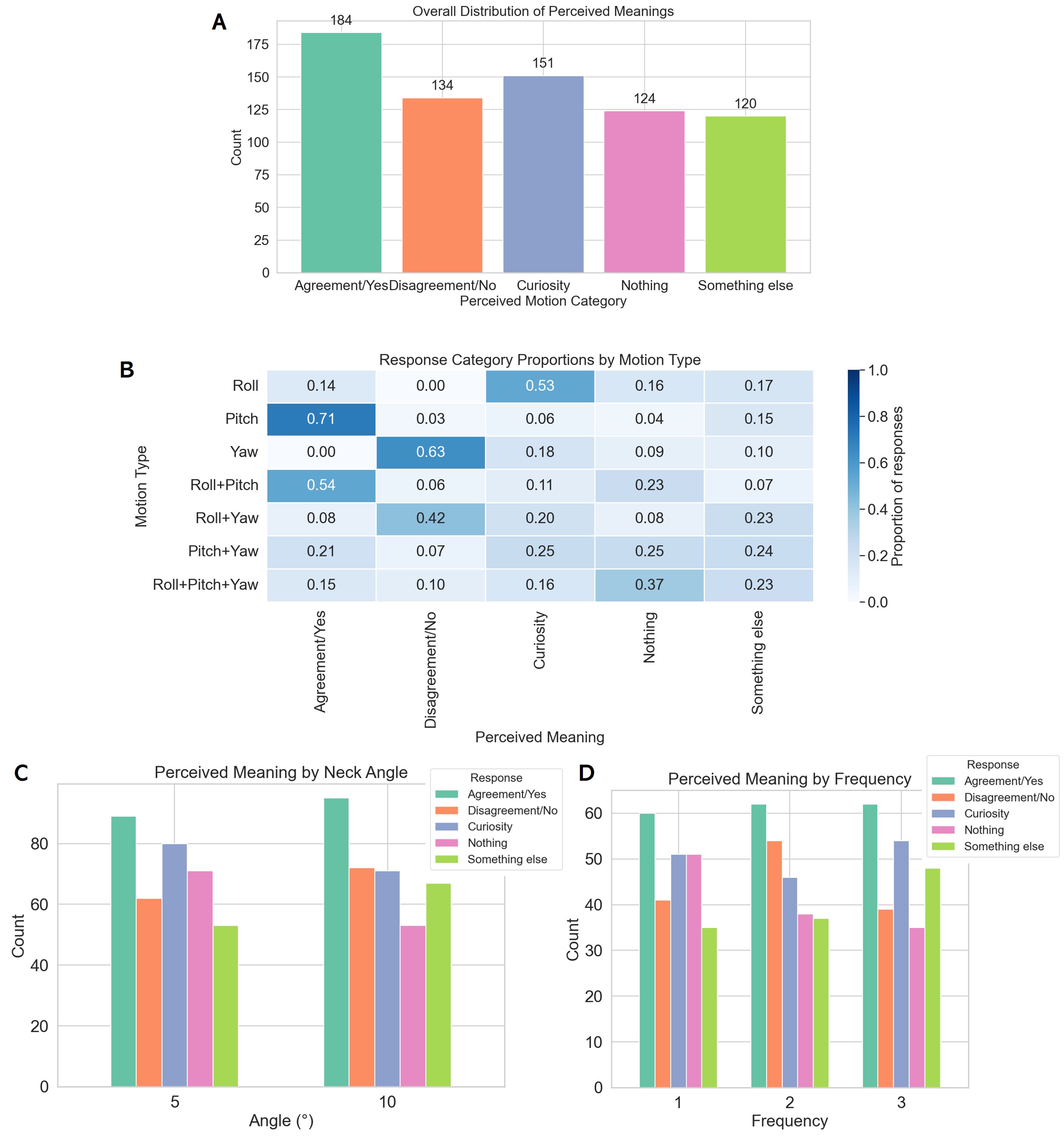}
    \caption{ Trends observed during the survey A) Overall distribution of perceived meanings across all 714 responses, showing the frequency with which participants selected each category B) Heatmap of mean perceived intensity as a function of motion type C)  Frequency of perceived meaning categories at each neck angle (5° vs. 10°), showing that larger angles reduced Nothing responses (from 20.0\% to 14.8\%), suggesting wider movements are more likely to convey a meaningful signal. D) Mean perceived intensity (±SE) by movement repetition frequency. }
    \label{survey1}
\end{figure}

The trends observed during the survey are shown in Figure \ref{survey1}. 
Across all stimuli, Agreement/Yes was the most frequently perceived meaning (25.8\%), followed by Curiosity (21.1\%), Disagreement/No (18.8\%), Nothing (17.4\%), and Something else (16.8\%). The mean perceived intensity across all stimuli was 2.37 out of 5 (SD = 1.20). This can be seen in Figure \ref{survey1}A. 
The axis of rotation had a strong and consistent influence on how the robot's head motion was interpreted (\ref{survey1}B). Pure Pitch motion was most strongly associated with Agreement/Yes (71.4\% of responses), with no participant interpreting it as Disagreement/No (0.0\%), confirming that vertical nodding is a highly unambiguous affirmative signal. Pure Yaw motion was most strongly associated with Disagreement/No (63.4\%), and was never interpreted as Agreement/Yes (0.0\%), consistent with the conventional head-shake for negation. Pure Roll motion was predominantly read as Curiosity (53.5\%). Combined Roll+Pitch motion preserved an affirmative leaning (Agreement/Yes, 53.8\%), while Roll+Yaw shifted toward Disagreement/No (41.9\%). The most perceptually ambiguous stimuli were those combining all three axes (Roll+Pitch+Yaw), where Nothing was the plurality response (37.3\%), and Pitch+Yaw, where responses were almost evenly split across Curiosity (24.5\%), Nothing (24.5\%), and Something else (23.5\%), indicating that compound multi-axis motions substantially reduce interpretive clarity. A Kruskal–Wallis test confirmed that axis combination had a statistically significant effect on perceived intensity (H = 16.71, p = 0.010), with Pitch-only motions yielding the highest mean intensity score (M = 2.80, SD = 1.12), and Roll+Yaw the lowest (M = 2.18, SD = 1.04).

Neck angle had a significant and positive effect on perceived intensity (Figure \ref{survey1}B). Motions executed at 10° produced substantially higher intensity ratings (M = 2.57, SD = 1.18) than those at 5° (M = 2.16, SD = 1.19), a difference confirmed as statistically significant by both Kruskal–Wallis (H = 21.73, p < 0.001) and Spearman correlation (r = 0.178, p < 0.001). Larger angles also shifted categorical perception, with Nothing responses dropping from 20.0\% at 5° to 14.8\% at 10$^\circ$, suggesting that wider movements are more likely to convey a meaningful signal. 
Repetition frequency (1–3 cycles) showed a weak but statistically significant positive correlation with intensity (r = 0.083, p = 0.029), with mean intensity increasing monotonically from 2.24 (1 repetition) to 2.41 (2 repetitions) to 2.47 (3 repetitions), though the overall Kruskal–Wallis test did not reach significance (H = 4.92, p = 0.085), suggesting frequency has only a marginal influence on perceived intensity. This is illustrated in Figure \ref{survey1}D.

\begin{figure}[!b]
    \centering
    \includegraphics[width=0.97\linewidth]{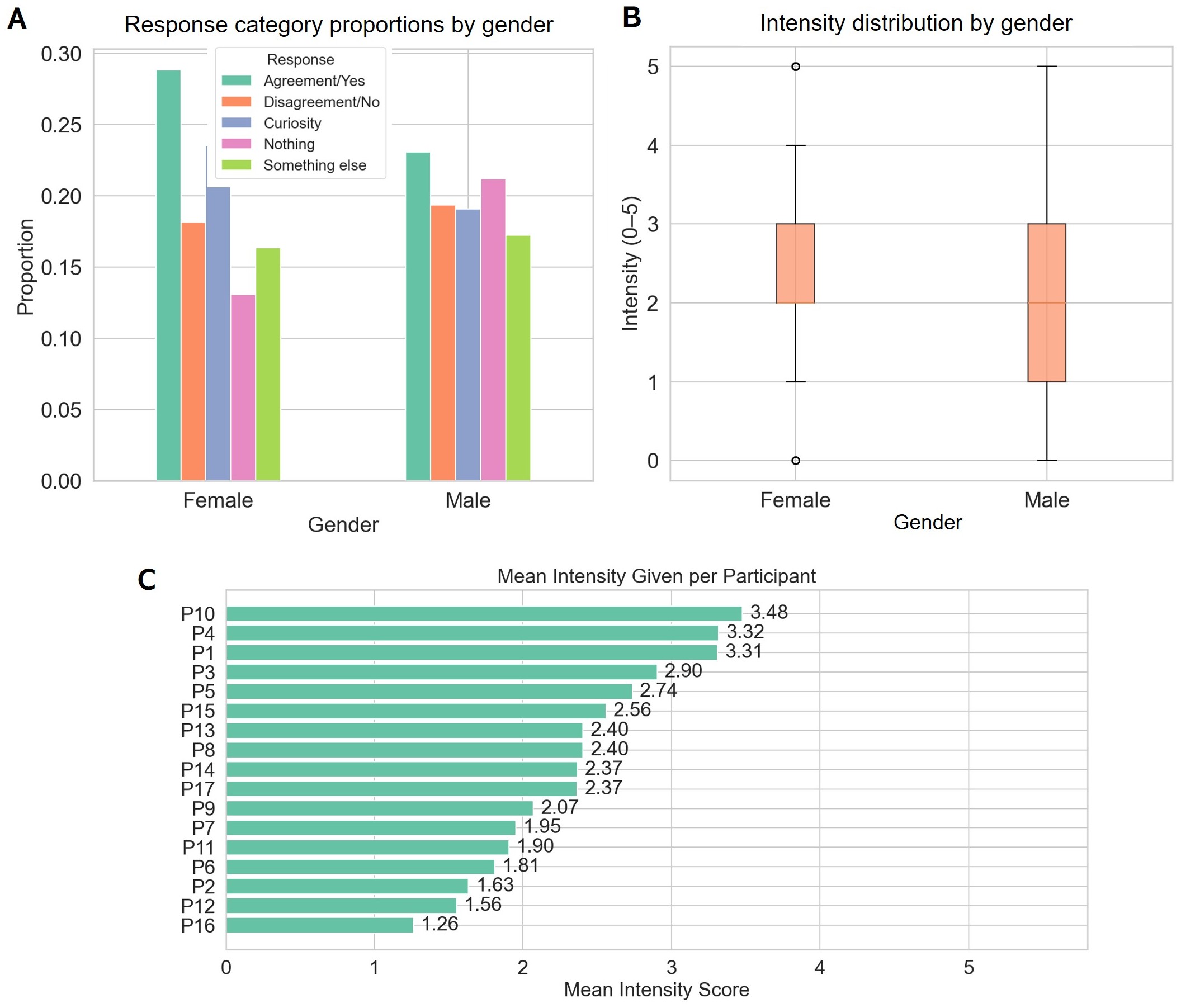}
    \caption{Gender differences in perception A) proportion of perceived meaning categories for female and male participants, showing females more frequently selected Agreement/Yes (28.9\% vs. 23.1\%) and males more frequently reported Nothing (21.2\% vs. 13.1\%) B)  boxplot of intensity scores by gender, with females assigning marginally higher ratings (M = 2.46 vs. 2.29), though the difference was not statistically significant (H = 3.27, p = 0.071). C)  Mean perceived intensity (±SE) by movement repetition frequency.}
    \label{survey2}
\end{figure}

Results related to gender-based trends observed during the survey are shown in Figure \ref{survey2}. Female participants assigned slightly higher intensity scores on average (M = 2.46, SD = 1.16) compared to male participants (M = 2.29, SD = 1.24), though this difference did not reach statistical significance with 17 participants (H = 3.27, p = 0.071). This can be seen in Figure \ref{survey2}A and B. Female participants showed a stronger tendency to interpret movements as Agreement/Yes (28.9\% vs. 23.1\%) and Curiosity (23.5\% vs. 19.1\%), while male participants more frequently reported Nothing (21.2\% vs. 13.1\%), suggesting a possible gender-linked difference in sensitivity to robotic non-verbal cues that warrants further investigation with a larger sample.

Across the 84 stimuli, the mean proportion of participants selecting the most-chosen response category was 0.546, indicating moderate overall agreement. Fifty-seven of the 84 stimuli (67.9\%) achieved majority agreement ($\geq$50\%), while 16 stimuli (19.0\%) reached high consensus ($\geq$75\%). The highest agreement observed for any single stimulus was 88.9\%, and the lowest was 33.3\%, highlighting considerable variation in the perceptual clarity of different motion designs. Figure \ref{survey2}C suggests that variations in individual characteristics and backgrounds may contribute to differences in perception. While these results support the validity of our findings, further in-depth studies are needed to confirm the observed trends, as gender, cultural and demographic factors may underlie the responses.

\section{Discussions}

The gap between 6.97 bits (geometric) and H (empirical from pixels)  tells us how much information is lost due to visual ambiguity, which directly relates to how well an observer could interpret the robot's communicative intent.

The most counterintuitive and practically important finding is that 3 DoF produces less communicative information than 2 DoF despite consuming significantly more energy. For robot designers, this challenges the default assumption that adding mechanical complexity improves expressiveness. A humanoid robot with a full roll-pitch-yaw neck is not automatically a better communicator than one with two axes and it is a more expensive and less efficient one. The design principle that follows is: stop at 2 DoF for neck-based nonverbal communication unless there is a specific task requirement for the third axis.

The Pareto frontier analysis shows that the roll-yaw combination at $10^\circ$, 0.5 m/s$^2$, 3 Hz delivers the highest communicative information at competitive energy cost. This is a 2 DoF configuration that omits pitch which is the most anatomically prominent neck motion in humans (nodding). For robot designers this is significant: pitch is mechanically expensive and informationally redundant when combined with other axes. A robot neck that prioritises yaw and roll over pitch may be a better communicative platform than one that mimics the full human range of motion.

The negative relationship between acceleration and entropy is a direct design guideline. Robots that move their necks quickly and abruptly  which might seem more dynamic and expressive. This actually produce less informationally rich pixel signals.  For communication-oriented robots, motion planning should favour low acceleration profiles regardless of DoF configuration. This applies equally to humanoids, non-humanoid service robots, and any platform where neck or head motion is used to signal attention, intent, or emotional state.

Yaw has the lowest energy cost of all single axes and maintains competitive entropy. For mobile robots, wearable robots, or any platform with a strict power budget where a full articulated neck is impractical, a single yaw-axis head motion may deliver the best communicative return per joule.  This has direct implications for lightweight social robots, drone-mounted heads, and prosthetic or assistive devices where energy efficiency is critical.

The Motor Information Space framework is not specific to necks. Any robot body part that produces observable motion, for instance an arm, a torso, a wheeled base rotating in place, can in principle be mapped onto the same H vs E space.  The ENIE metric and Pareto frontier analysis provide a general methodology for evaluating the communicative efficiency of any robot motion modality. A robot without a neck can use these principles to optimise the communicative content of its available degrees of freedom, whether that is wrist rotation, shoulder tilt, or whole-body orientation.

The best individual condition (roll+yaw, $10^\circ$ , 0.5 m/s², 3 Hz at 5.258 bits) significantly outperforms the mean 2 DoF condition, and even outperforms the mean 3 DoF condition. This means how a robot moves its neck matters as much as what neck it has. For existing robots with fixed morphologies, motion parameter optimisation, eg: angle, frequency, and especially acceleration, is a low-cost route to substantially improving communicative effectiveness without hardware changes. The simple lattice mechanism used to create neck motions in this study testify to this.

The divergence of entropy and energy at 3 DoF is likely not unique to robot necks. Here, adding mechanical complexity yields diminishing informational return while energy continues to rise. It may represent a general phenomenon in robotic morphology: there exists for any body part a complexity threshold beyond which additional degrees of freedom produce redundant rather than novel motion patterns. Identifying this threshold for other body parts such as hands, arms, torsos, using the same information-theoretic methodology could become a systematic approach to biomechanically-informed robot design that does not simply copy human anatomy but instead identifies the minimum sufficient morphology for a given communicative function. In addition, paying attention to nonverbal gestures similar to the neck motions discussed, will help robots sunchronize with their human counterparts during interaction \cite{hu2024human}.

For humanoid robots specifically, these results suggest that the neck should not be treated as a passive structural component or a direct copy of human cervical anatomy. Instead, neck morphology should be selected based on the communicative requirements of the deployment context. A humanoid designed for social interaction in a hospital or care setting, should prioritise a 2 DoF roll-yaw configuration operating at large angles and low acceleration. This is  where clear, readable signals of attention and acknowledgement matter. A humanoid designed for rapid task execution in an industrial setting may not need a communicative neck at all, saving weight, energy, and mechanical complexity for more task-critical components.

\subsection{Limitations}
The robot had only one gender representation, which limited the diversity and realism of human-like interaction during gesture demonstrations. Furthermore the robot was slower in performing the gestures compared to humans, which may have affected the naturalness and timing of motion analysis. The range of neck motion in the robot was more restricted than in humans, reducing the accuracy of replicating natural neck movements.
Mechanical movement lacked the subtle muscle dynamics and smooth transitions present in human neck motion. Future work will look at improving these aspects of the development of robotic necks.

\section{Conclusions}

This study demonstrates that communicative effectiveness in robot neck motion is governed by the interaction between morphological configuration, motion parameters, and energy expenditure. The central finding is that 3 DoF produces less Shannon entropy than 2 DoF despite consuming significantly more energy. This directly challenges the assumption that additional degrees of freedom improve expressiveness. Beyond a complexity threshold, simultaneous multi-axis activation produces redundant visual patterns, reducing informational richness: the morphological information bottleneck. For robot designers, a 2 DoF roll-yaw configuration at large angles, low acceleration, and high frequency represents the optimal balance between communicative capacity and energy efficiency, lying on the Pareto frontier of the Motor Information Space. For energy-constrained platforms, a single yaw axis delivers the best communicative return per joule. Pitch, despite being the most prominent human neck movement, is mechanically expensive and informationally redundant in combination with other axes.
Parameter choice rivals morphology in determining communicative output. Slow deliberate movements consistently outperform fast aggressive ones, meaning motion planning is as consequential as hardware selection for communication-oriented robots. Beyond neck design, the Motor Information Space and ENIE metric are general tools applicable to any robot motion modality, providing a principled quantitative basis for identifying the minimum sufficient morphology for communicative function across broader human-robot interaction platforms. In addition, it provided insight into artificial communication niches by combining human capabilities, such as multiple degrees of freedom, with artificial and augmented features to support the development of cyborg systems.

\medskip
\textbf{Data Availability} \par 
Video dataset for robotic neck movements discussed in the study can be found \href{https://zenodo.org/records/20577859}{here}. 

\medskip
\textbf{Ethics statement} \par 
Survey in this study was approved by the Human Research Ethics Committee (HREC) of École Polytechnique Fédérale de Lausanne (EPFL) under approval number HREC000763. 

\medskip
\newpage
\textbf{Conflict of Interest} \par 
Authors declare no conflict of interest.

\medskip
\textbf{Acknowledgements} \par 
The authors would like to thank the volunteers who contributed to the analysis. This work was partially supported by Horizon 2020 research and innovation programme under the Marie Skłodowska-Curie grant agreement No 101034337. The work was performed while the corresponding author was associated with the University of Cambridge. The authors used generative AI tools to improve the language and grammar of this manuscript. All suggested edits were reviewed and approved by the authors.

\medskip

%
\bibliographystyle{unsrt}
\bibliography{References}

\newpage
\section{Supplementary Information}

\subsection{Further details related to motor-based analysis}
\label{sup_energy}

\subsection*{Signal Acquisition}

At each time step $k$, two six-dimensional vectors were read from the
controller:
 
\begin{itemize}
  \item $\mathbf{I}(k) = \bigl[I_0,\, I_1,\, I_2,\, I_3,\, I_4,\, I_5\bigr]^\top$\;[A]
        — measured joint motor currents, obtained via \texttt{getActualCurrent()}.
  \item $\boldsymbol{\omega}(k) = \bigl[\omega_0,\, \omega_1,\, \omega_2,\, \omega_3,\, \omega_4,\, \omega_5\bigr]^\top$\;[rad \, s$^{-1}$]
        — measured joint velocities, obtained by \texttt{getActualQd()}.
\end{itemize}
 
The sampling interval was fixed at $\Delta t = 1/125 = 8\,\text{ms}$,
matching the UR3 real-time control loop. A software timer consumed the
remainder of each slot to maintain consistent cadence:
 
\begin{lstlisting}[caption={125\,Hz timing loop.}]
elapsed   = time.time() - t_sample
remaining = SAMPLE_DT - elapsed      # SAMPLE_DT = 1 / 125
if remaining > 0:
    time.sleep(remaining)
\end{lstlisting}
 
\subsection*{Torque Constants}

\begin{table}[h]
  \centering
  \caption{UR3 motor torque constants used in the study.}
  \label{tab:kt}
  \begin{tabular}{llcc}
    \toprule
    Joint & Description  & $K_{t,i}$ [Nm\,A$^{-1}$] \\
    \midrule
    $i=0$ & Base         & 0.1249 \\
    $i=1$ & Shoulder     & 0.1249 \\
    $i=2$ & Elbow        & 0.0934 \\
    $i=3$ & Wrist\,1     & 0.0453 \\
    $i=4$ & Wrist\,2     & 0.0453 \\
    $i=5$ & Wrist\,3     & 0.0453 \\
    \bottomrule
  \end{tabular}
\end{table}

\subsection*{Energy Integration}
 
\begin{lstlisting}[
  language=Python,
  caption={Trapezoidal integration with actual elapsed time.}
]
delta_e = 0.5 * (state["prev_power"].sum() + powers.sum()) * dt
state["total_energy"] += delta_e
state["prev_power"]    = powers.copy()
\end{lstlisting}
 
\subsection*{Motion Execution and Concurrent Logging}
 
The trajectory consisted of $N$ oscillation cycles on selected joints,
each cycle comprising a negative displacement ($-\theta$) followed by a
positive displacement ($+\theta$) from the home configuration, and a
final return to home. Each sub-move was commanded via the RTDE control
interface:
 
\begin{lstlisting}[caption={Non-blocking joint move with concurrent sampling.}]
rtde_c.moveJ(target, SPEED, ACCEL, asynchronous=True)
log_loop(rtde_r, logger, phase, duration, state)
\end{lstlisting}
 
Setting \texttt{asynchronous=True} returns control to the Python process
immediately, allowing \texttt{log\_loop()} to acquire samples throughout the
motion. The logging duration for each sub-move was estimated from the
trapezoidal velocity profile:
 
\begin{equation}
  T_{\text{est}} =
  \begin{cases}
    \dfrac{d}{v_{\max}} + \dfrac{v_{\max}}{a_{\max}} + T_m
      & \text{if } \dfrac{v_{\max}^2}{a_{\max}} \leq d, \\[10pt]
    2\sqrt{\dfrac{d}{a_{\max}}} + T_m
      & \text{otherwise,}
  \end{cases}
  \label{eq:dest}
\end{equation}
 
where $d$ is the maximum joint displacement among all joints,
$v_{\max}$ and $a_{\max}$ are the commanded speed and acceleration,
and $T_m = 0.8\,\text{s}$ is a fixed settling margin appended after
the estimated motion duration.
 
\subsection*{Phase Tagging and Data Recording}
 
Each sample was labelled with a phase identifier — \texttt{baseline},
\texttt{cycle\,$n$\_minus}, \texttt{cycle\,$n$\_plus}, or
\texttt{return\_home} — enabling per-phase energy breakdowns in
post-processing. Every sample was appended in real time to a CSV file
containing the fields listed in Table~\ref{tab:csv}.
 
\begin{table}[h]
  \centering
  \caption{Fields written to the per-sample log file
           (\texttt{ur3\_torque\_log\_<timestamp>.csv}).}
  \label{tab:csv}
  \begin{tabular}{lll}
    \toprule
    Field & Unit & Description \\
    \midrule
    \texttt{timestamp\_s}         & s    & Wall-clock time of sample \\
    \texttt{phase}                & —    & Trajectory phase label \\
    \texttt{current\_j$i$\_A}     & A    & $I_i(k)$, $i \in \{0,\ldots,5\}$ \\
    \texttt{torque\_j$i$\_Nm}     & Nm   & $\tau_i(k)$ (Eq.~\ref{eq:torque}) \\
    \texttt{power\_j$i$\_W}       & W    & $P_i(k)$ (Eq.~\ref{eq:power}) \\
    \texttt{total\_power\_W}      & W    & $P_{\text{total}}(k)$ (Eq.~\ref{eq:ptotal}) \\
    \texttt{cumulative\_energy\_J}& J    & $E(k)$ (Eq.~\ref{eq:energy}) \\
    \bottomrule
  \end{tabular}
\end{table}
 
A separate summary file (\texttt{ur3\_energy\_summary\_<timestamp>.csv})
was written at the end of each run, recording the total integrated energy
$E_{\text{total}}$, peak and mean power, and per-joint peak torque and
current values.
 
\subsection*{Timing Integrity}
 
The \texttt{prev\_time} variable used to compute $\Delta t_k$ in
Eq.~\eqref{eq:energy} was reset locally at the start of each call to~\texttt{log\_loop()}, rather than inheriting a module-level timestamp.
This prevents a spurious large $\Delta t$ on the first sample of each
phase — which would otherwise arise from the wall-clock time elapsed during
robot connection and trajectory planning — from inflating the energy estimate.
 
 
Table~\ref{tab:decisions} summarises the four implementation choices that
distinguish this approach from a naive implementation and their physical
justification.
 
\begin{table}[h]
  \centering
  \caption{Key implementation decisions and rationale.}
  \label{tab:decisions}
  \begin{tabular}{p{3.8cm} p{5.0cm} p{5.0cm}}
    \toprule
    Decision & Naive approach & This work \\
    \midrule
    Torque model
      & $\tau_i = K_{t,i} \cdot G \cdot I_i$ (gear ratio applied separately)
      & $\tau_i = K_{t,i} \cdot I_i$; $K_{t,i}$ already encodes gearbox \\[6pt]
    Power sign
      & $P_i = \tau_i \cdot \omega_i$ (signed; braking cancels)
      & $P_i = |\tau_i \cdot \omega_i|$ (absolute; no cancellation) \\[6pt]
    Motion logging
      & Blocking \texttt{moveJ}; samples taken only after motion ends
      & \texttt{asynchronous=True}; samples taken throughout motion \\[6pt]
    Time-step $\Delta t$
      & Fixed nominal; \texttt{prev\_time} set at module load
      & Actual elapsed time; \texttt{prev\_time} reset per phase \\
    \bottomrule
  \end{tabular}
\end{table}

\subsubsection{Plot 3A — Total Energy Consumption}

Total electrical energy $E$ computed according to 
Eq.~\ref{eq:energy} is plotted against motion type for both 
angle conditions $\theta \in \{5^{\circ}, 10^{\circ}\}$. 
Each marker represents the mean energy across repeated trials 
for a given motion type and angle. A polynomial trendline is 
fitted separately for each angle:

\begin{equation}
    \hat{E}(x) = \sum_{k=0}^{d} \alpha_k x^k, \quad d = 2
\end{equation}

\noindent where $x$ is the integer-coded motion type index and 
$\alpha_k$ are the fitted polynomial coefficients obtained via 
least squares. Solid line corresponds to $\theta = 5^{\circ}$ 
and dashed line to $\theta = 10^{\circ}$.

\subsubsection{Plot 3B — Peak Power Consumption}

Peak power $P_{\text{peak}}$ defined in Eq.~\ref{eq:peak_power} 
is plotted against motion type for both angle conditions. The 
same polynomial trendline fitting procedure as Plot A is applied:

\begin{equation}
    \hat{P}_{\text{peak}}(x) = \sum_{k=0}^{d} \beta_k x^k, 
    \quad d = 2
\end{equation}

\noindent Peak power captures the maximum instantaneous 
loading on the robot actuators and is particularly relevant 
for assessing hardware stress under combined multi-axis 
movements.

\subsubsection{Plot 3C — Mean Power Consumption}

Mean power $\bar{P}$ defined in Eq.~\ref{eq:mean_power} is 
plotted against motion type for both angle conditions, using 
the same trendline procedure:

\begin{equation}
    \hat{\bar{P}}(x) = \sum_{k=0}^{d} \gamma_k x^k, \quad d = 2
\end{equation}

\noindent Mean power reflects the sustained electrical load 
during movement and complements peak power by characterising 
the average rather than worst-case energy demand.

\subsubsection{Plot 3D — Energy per Degree of Freedom}

To assess whether energy scales proportionally with the number 
of active axes, total energy is normalised by the DoF count:

\begin{equation}
    E_{\text{DoF}} = \frac{E}{n_{\text{DoF}}}
    \label{eq:energy_DoF}
\end{equation}

\noindent where $n_{\text{DoF}} \in \{1, 2, 3\}$ is the number 
of simultaneously active rotation axes. A flat trend across 
motion types indicates proportional energy scaling, while a 
rising trend indicates super-linear interaction costs between 
axes. Both angle conditions are shown with the same polynomial 
trendline fitting procedure as Plots A--C.

\subsubsection{Plot 3E — Delta Energy Between $10^\circ$ 
and $5^\circ$ Movements}

The additional energy cost incurred by increasing the movement 
amplitude from $5^{\circ}$ to $10^{\circ}$ is computed per 
motion type as:

\begin{equation}
    \Delta E = E_{10^{\circ}} - E_{5^{\circ}}
    \label{eq:delta_energy}
\end{equation}

\subsection{Pixel-based Analysis}
\label{Pixel_plots}

All plots use Shannon entropy $H$ as the dependent variable, computed 
from the pixel-change signal $\delta_t$ extracted from 84 video 
recordings of a UR3 robot neck across combinations of roll, pitch, 
and yaw axes. The pixel-change signal is defined as in eq. \ref{pixel}. Shannon entropy is computed as in eq. \ref{H}.  

\noindent\textbf{Key parameters:} $B = 64$ histogram bins, 
logarithm base 2 (bits), fixed camera, static background.

\noindent\textbf{Key assumptions:} All pixel change originates 
from neck movement only. The pixel-change distribution is stationary 
over each video. Higher $H$ indicates richer communicative content.

Experimental parameter space is shown in Figure \ref{exp}A.  Considered axes include roll, pitch, yaw and all combinations (1, 2, 3 DoF), angles  $\theta \in \{5^{\circ}, 10^{\circ}\}$, accelerations $a \in \{0.5, 2.0\}$ m/s$^2$, frequencies  $f \in \{1, 2, 3\}$ Hz across all 84 videos ($N = 84$).

\subsubsection{Figure \ref{Pixel_plot_2}A — Acceleration vs Pixel Entropy}

Mean $H$ is plotted against $a \in \{0.5, 2.0\}$ m/s$^2$ with 
separate lines per DoF level. Each point represents the mean entropy 
across all videos sharing the same acceleration and DoF, with 
$\pm 1$ standard deviation bands:

\begin{equation}
    \bar{H}_{a,\,\text{DoF}} = \frac{1}{|\mathcal{S}|} 
    \sum_{k \in \mathcal{S}} H_k
\end{equation}

\noindent where $\mathcal{S}$ is the subset of videos with a given 
acceleration and DoF level.

\subsubsection{Figure \ref{Pixel_plot_2}B — DoF vs Pixel Entropy}

Mean $H$ is plotted against DoF $\in \{1, 2, 3\}$ with separate 
lines per angle level $\theta \in \{5^{\circ}, 10^{\circ}\}$, 
averaged across all acceleration and frequency conditions:

\begin{equation}
    \bar{H}_{\text{DoF},\,\theta} = \frac{1}{|\mathcal{S}|} 
    \sum_{k \in \mathcal{S}} H_k
\end{equation}

\noindent where $\mathcal{S}$ is the subset of videos with a given 
DoF and angle.

\subsubsection{Figure \ref{Pixel_plot_2}C — Frequency vs Pixel Entropy}

Mean $H$ is plotted against $f \in \{1, 2, 3\}$ Hz with separate 
lines per DoF level, averaged across all angle and acceleration 
conditions:

\begin{equation}
    \bar{H}_{f,\,\text{DoF}} = \frac{1}{|\mathcal{S}|} 
    \sum_{k \in \mathcal{S}} H_k
\end{equation}

\subsubsection{Figure \ref{Pixel_plot_2}D — Angle vs Pixel Entropy}

Mean $H$ is plotted against $\theta \in \{5^{\circ}, 10^{\circ}\}$ 
with separate lines per DoF level, averaged across all acceleration 
and frequency conditions:

\begin{equation}
    \bar{H}_{\theta,\,\text{DoF}} = \frac{1}{|\mathcal{S}|} 
    \sum_{k \in \mathcal{S}} H_k
\end{equation}

\subsubsection{Figure \ref{Pixel_plot_2}E — Entropy Distribution per DoF}

A boxplot of the full distribution of $H$ values is shown per DoF 
level, aggregated across all $\theta$, $a$, and $f$ conditions. 
The box spans the interquartile range (Q1 to Q3), the median line 
represents Q2, and whiskers extend to 1.5 times the IQR. Individual 
data points are overlaid as a strip plot:

\begin{equation}
    \text{IQR}_{\text{DoF}} = Q3_{\text{DoF}} - Q1_{\text{DoF}}
\end{equation}

\subsubsection{Figure \ref{Pixel_plot_2}F — ENIE vs DoF}

Error bars represent $\pm 1$ standard deviation 
across conditions within each DoF level. Key values: 
$\text{ENIE}_{1\,\text{DoF}} = 0.0121$, 
$\text{ENIE}_{2\,\text{DoF}} = 0.0061$, 
$\text{ENIE}_{3\,\text{DoF}} = 0.0039$ bits/J. 
This metric assumes that electrical energy is a valid proxy for 
mechanical cost and that energy scales with the number of active DoF.

\subsubsection{Figure \ref{Pixel_plot_2}G — Angle $\times$ Acceleration Heatmap}

Mean entropy is computed for each cell of the $2 \times 2$ grid 
defined by $\theta \in \{5^{\circ}, 10^{\circ}\}$ and 
$a \in \{0.5, 2.0\}$ m/s$^2$, averaged across all DoF and 
frequency conditions:

\begin{equation}
    \bar{H}_{\theta,\, a} = \frac{1}{|\mathcal{S}|} 
    \sum_{k \in \mathcal{S}} H_k
\end{equation}

\noindent Key values: $\bar{H}_{10^{\circ},\, 0.5} = 5.01$ bits 
(maximum), $\bar{H}_{5^{\circ},\, 2.0} = 4.54$ bits (minimum).

\subsubsection{Figure \ref{Pixel_plot_2}H — DoF $\times$ Frequency Heatmap}

Mean entropy is computed for each cell of the $3 \times 3$ grid 
defined by DoF $\in \{1, 2, 3\}$ and $f \in \{1, 2, 3\}$ Hz, 
averaged across all angle and acceleration conditions:

\begin{equation}
    \bar{H}_{\text{DoF},\, f} = \frac{1}{|\mathcal{S}|} 
    \sum_{k \in \mathcal{S}} H_k
\end{equation}

\noindent Maximum cell: $\bar{H}_{2\,\text{DoF},\, 3\,\text{Hz}}= 5.00$ bits. Minimum cell: $\bar{H}_{3\,\text{DoF},\, 1\,\text{Hz}} = 4.43$ bits.

\end{document}